\documentclass[conference]{IEEEtran}
\IEEEoverridecommandlockouts
\usepackage{cite}
\usepackage{amsmath,amssymb,amsfonts}
\usepackage{algorithmic}
\usepackage{graphicx}
\usepackage{textcomp}
\usepackage{xcolor}
\usepackage{hyperref}
\usepackage{siunitx}
\def\BibTeX{{\rm B\kern-.05em{\sc i\kern-.025em b}\kern-.08em T\kern-.1667em\lower.7ex\hbox{E}\kern-.125emX}}
\usepackage{tikz}
\usetikzlibrary{positioning, shapes.multipart}
\usepackage{booktabs}
\usepackage{tabularx}     
\newcolumntype{Y}{>{\centering\arraybackslash}X}
\usepackage{float}        
\usepackage{array}
\usepackage{dblfloatfix}
\usepackage[utf8]{inputenc}   
\usepackage{multirow}
\usepackage{newunicodechar}
\newunicodechar{≈}{\approx}

\DeclareUnicodeCharacter{202F}{~}
\begin{document}

\title{BEAVR: Bimanual, multi-Embodiment, Accessible, Virtual Reality Teleoperation System for Robots}

\author{
\IEEEauthorblockN{Alejandro Posadas-Nava* \thanks{*Equal contribution}}
\IEEEauthorblockA{\textit{Department of Aeronautics and Astronautics} \\
Massachusetts Institute of Technology, Cambridge, USA \\
email or ORCID}
\and
\IEEEauthorblockN{Alejandro Carrasco*}
\IEEEauthorblockA{\textit{Department of Aeronautics and Astronautics} \\
Massachusetts Institute of Technology, Cambridge, USA \\
acarra@mit.edu}
\and
\IEEEauthorblockN{Richard Linares}
\IEEEauthorblockA{\textit{Department of Aeronautics and Astronautics} \\
Massachusetts Institute of Technology, Cambridge, USA \\
email or ORCID}
}

\maketitle

\begin{abstract}
\textbf{BEAVR} is an open-source, bimanual, multi-embodiment Virtual Reality (VR) teleoperation system for robots, designed to unify real-time control, data recording, and policy learning across heterogeneous robotic platforms. BEAVR enables real-time, dexterous teleoperation using commodity VR hardware, supports modular integration with robots ranging from 7-DoF manipulators to full-body humanoids, and records synchronized multi-modal demonstrations directly in the LeRobot dataset schema. Our system features a zero-copy streaming architecture achieving $\leq$35\,ms latency, an asynchronous ``think--act'' control loop for scalable inference, and a flexible network API optimized for real-time, multi-robot operation. We benchmark BEAVR across diverse manipulation tasks and demonstrate its compatibility with leading visuomotor policies such as ACT, DiffusionPolicy, and SmolVLA. All code is publicly available, and datasets are released on Hugging Face\footnote{Code, datasets, and VR app available at \url{https://github.com/ARCLab-MIT/BEAVR-Bot}.}
\end{abstract}


\section{Introduction}

Robots are transitioning from structured industrial lines to dynamic, human-centric settings, where \emph{dexterous, bimanual manipulation} is a prerequisite for useful autonomy. However, current teleoperation pipelines remain fragmented, often tied to specific hardware, proprietary formats, or custom frameworks, hindering reproducibility and slowing progress in human-robot interaction (HRI) and robotics research in general.

Teleoperation exists in various forms. Classical actuation typically involves transmitting forces from grounded handles to robot grippers. Leader-follower approaches mirror movements between two identical robotic arms, offering direct motion mapping but limiting versatility due to symmetry constraints and limited ability to mimic natural, diverse actions. More recently, Virtual Reality (VR)-based teleoperation offers a promising path forward. The availability of consumer-grade devices such as the Meta Quest has inspired recent systems such as OpenTeach \cite{iyer2024openteachversatileteleoperation} and Bunny VisionPro \cite{ding2024bunnyvisionprorealtimebimanualdexterous}, which enable bimanual demonstrations via immersive interfaces. Yet, these platforms are limited in key ways: they feature a steep learning curve, lack accessible system documentation, log data in non-standard formats, and offer no standardized bridge to learning-ready datasets.

We present \textbf{BEAVR}, an open-source, end-to-end VR teleoperation and policy learning system that unifies control, recording, and learning among heterogeneous robots. It enables real-time bimanual control of diverse end-effectors, such as a 7-degree-of-freedom (DoF) xArm or a full-body RX-1 humanoid, through a single VR interface, while capturing synchronized multi-modal demonstrations directly in the standardized LeRobot dataset format. The LeRobot framework \cite{cadene2024lerobot} provides a unified schema to store robot demonstrations along with a user-friendly policy learning pipeline, supporting a robotics community focused on making AI for real-world robotics more accessible. BEAVR also supports real-time RGB and RGB-D video streaming from commodity cameras and integrates precise tracking using Meta Quest 3S VR glasses. In particular, a complete setup that combines cameras, VR headset, and robotic hardware (RX-1 with LeapHand) can be assembled for approximately \$1000\footnote{Indicative bill of materials (July 2025): 16×Feetech STS3215‑C018 smart servos at \$34~ea \cite{feetech_servo} (total \$544), Meta-Quest-3S VR headset \$299 \cite{meta_quest3s}, LEAPHand parts kit \$200 \cite{shaw2023leaphandlowcostefficient}, and $\approx$1.5~kg PLA/TPU filament for printed frames $\approx$\$40 \cite{pla_filament}—yielding a total hardware cost of $\approx$\$1.08~k.}, underscoring the system’s affordability and potential for broad adoption. Beyond affordability, this work addresses a key question: \emph{can a single VR interface scale to multiple heterogeneous robots without incurring a latency penalty}?

\textbf{Contributions.} Our primary contributions are: (i) a hardware‑agnostic virtual-reality teleoperation interface that scales from a 7‑DoF tabletop arm to full‑body humanoids; (ii) a zero‑copy, latency‑aware streaming architecture that sustains sub‑35~ms end‑to‑end delay over commodity networks; (iii) a dataset‑native logger that exports demonstrations directly in the LeRobot schema for immediate use with modern imitation‑learning frameworks; and (iv) an open benchmark suite comprising six manipulation tasks, 50 expert demonstrations, and pretrained baseline policies.

The remainder of this paper is organized as follows. Section~\ref{sec:relatedWork} reviews prior teleoperation systems and datasets. Section~\ref{sec:system} details the BEAVR architecture. Section~\ref{sec:method} outlines the pipeline, including coordinate transformations, inverse kinematics, and networking. Section~\ref{sec:dataset} describes LeRobot integration and the dataset format. Section~\ref{sec:experiments} presents experimental results, including task success, policy learning, and system performance. Sections~\ref{sec:discussion} and~\ref{sec:limitationsandfutureWork} discuss findings, limitations, and future work. Section~\ref{sec:conclusion} concludes.

\section{Related Work}\label{sec:relatedWork}

Teleoperation has long been a key strategy for collecting reliable demonstrations across robotic tasks. Early systems typically relied on direct actuation methods, including leader-follower setups in which two identical robotic arms were physically or virtually linked to mirror each other’s motion. Other common approaches included joystick-based control or input devices such as 6-DoF controllers, which allowed users to command robot movement from a distance. These systems enabled early progress in remote manipulation, but often required hardware symmetry, offered limited dexterity, or depended on tightly coupled hardware-software configurations that were difficult to scale or adapt to new tasks.

Recent systems have explored VR and AR interfaces for more intuitive, human-centered teleoperation. These approaches use hand tracking \cite{iyer2024openteachversatileteleoperation, ding2024bunnyvisionprorealtimebimanualdexterous, guo2025telepreviewuserfriendlyteleoperationvirtual}, glove-based input \cite{zhang2025doglovedexterousmanipulationlowcost, fang2015noveldataglove}, or controller-based interaction \cite{imdieke2025sparkremotecosteffectiveremotebimanual} to control robot manipulators in real time. Some allow direct mapping of hand or arm motion to robot joints, while others introduce intermediate steps, such as moving a virtual end-effector in AR before executing commands on the physical robot\cite{vanhaastregt2024puppeteerrobotaugmentedreality}. The systems most similar to ours rely on VR hand tracking, sometimes augmented with haptic gloves or armbands. While effective for bimanual and finger-level manipulation, these platforms tend to be tied to specific hardware, are often not fully open source, and either lack affordability or store data in proprietary formats. Some systems address one or two of these limitations, but very few meet all of them together. Without openness, low cost, hardware flexibility, and standardized data output, it is difficult to achieve reusability, extensibility, or broad adoption.

Open-source projects like LEAP Hand \cite{shaw2023leaphandlowcostefficient} and RX-1 \cite{rx1_robot} showcase creative, low-cost integration, yet they still lack a standardized software ecosystem for broad reuse.

\section{System Overview}\label{sec:system}

BEAVR is an open-source teleoperation system that translates VR-based human hand movements into real-time robot control, enabling dexterous, multi-embodiment manipulation. The architecture is organized into three main modules: teleoperation, data collection, and model training.

The \textbf{teleoperation module} uses a modular component-based design. It consists of three core components that run as separate processes and communicate via lightweight \textsf{ZMQ} channels:
\begin{itemize}
    \item \textbf{Detector:} Captures raw data from the VR headset, including hand keypoints, button presses, and session commands.
    \item \textbf{Operator:} Processes and transforms the input data into robot-relevant commands, such as end-effector positions or joint targets.
    \item \textbf{Interface:} Sends the high-level actions to the robot’s control system, ensuring smooth and stable execution.
\end{itemize}
This modular design allows flexible integration of new hardware and robots, while fault isolation ensures that if one module fails, the others remain operational.

The \textbf{data collection module} interfaces seamlessly with the LeRobot framework~\cite{cadene2024lerobot}. It records synchronized streams of observations, actions, and metadata into a standardized, hardware-agnostic format, enabling immediate downstream use for benchmarking and learning.

The \textbf{model training module} leverages LeRobot’s training pipeline but enhances it with asynchronous inference. A dedicated control loop streams actions at a fixed rate, while a separate inference thread computes policy outputs as compute resources become available. This decoupled scheme preserves real-time performance, even under high computational load.

Overall, BEAVR combines modularity, real-time performance, and open data standards to provide a scalable platform for research in teleoperated, dexterous robot control.


\begin{figure}
    \centering
    \includegraphics[width=0.95\linewidth]{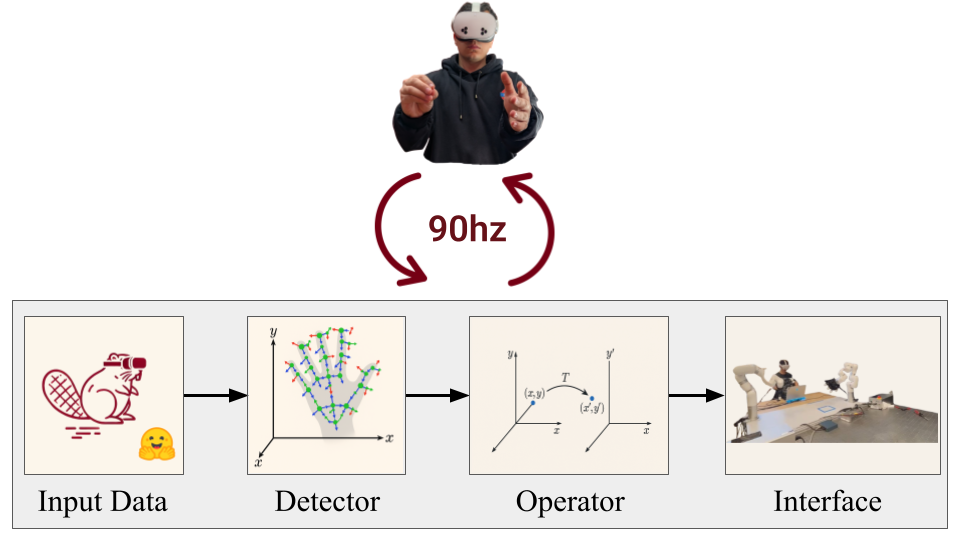}
    \caption{Data flow visualization through the system components of the teleoperation stack. Hand keypoint data flows from the input device (a VR headset) to the detector. The operator takes the raw keypoint data, creates a coordinate frame, computes coordinate transformations, and sends a transformation matrix to the interface, which commands the robot}
    \label{fig:components}
\end{figure}

\section{BEAVR}\label{sec:method}

BEAVR is a modular teleoperation pipeline designed to connect VR-based hand tracking to robot actuation across heterogeneous robotic platforms. The system is hardware-agnostic and has been deployed on a range of embodiments, including a 16-DoF dexterous robotic hand, a 7-DoF anthropomorphic arm, and an in-house assembled 6-DoF RX-1 robotic humanoid. For the RX-1 platform, BEAVR integrates seamlessly with ROS-based components, enabling direct interfacing with ROS control stacks and facilitating compatibility with existing robotic software ecosystems. This flexibility allows BEAVR to support diverse experimental setups, from fine-grained hand manipulation to whole-arm teleoperation, through dedicated modules for inverse kinematics (IK), temporal smoothing, and real-time control.

\begin{figure}[t]
    \centering
    \includegraphics[width=0.9\linewidth]{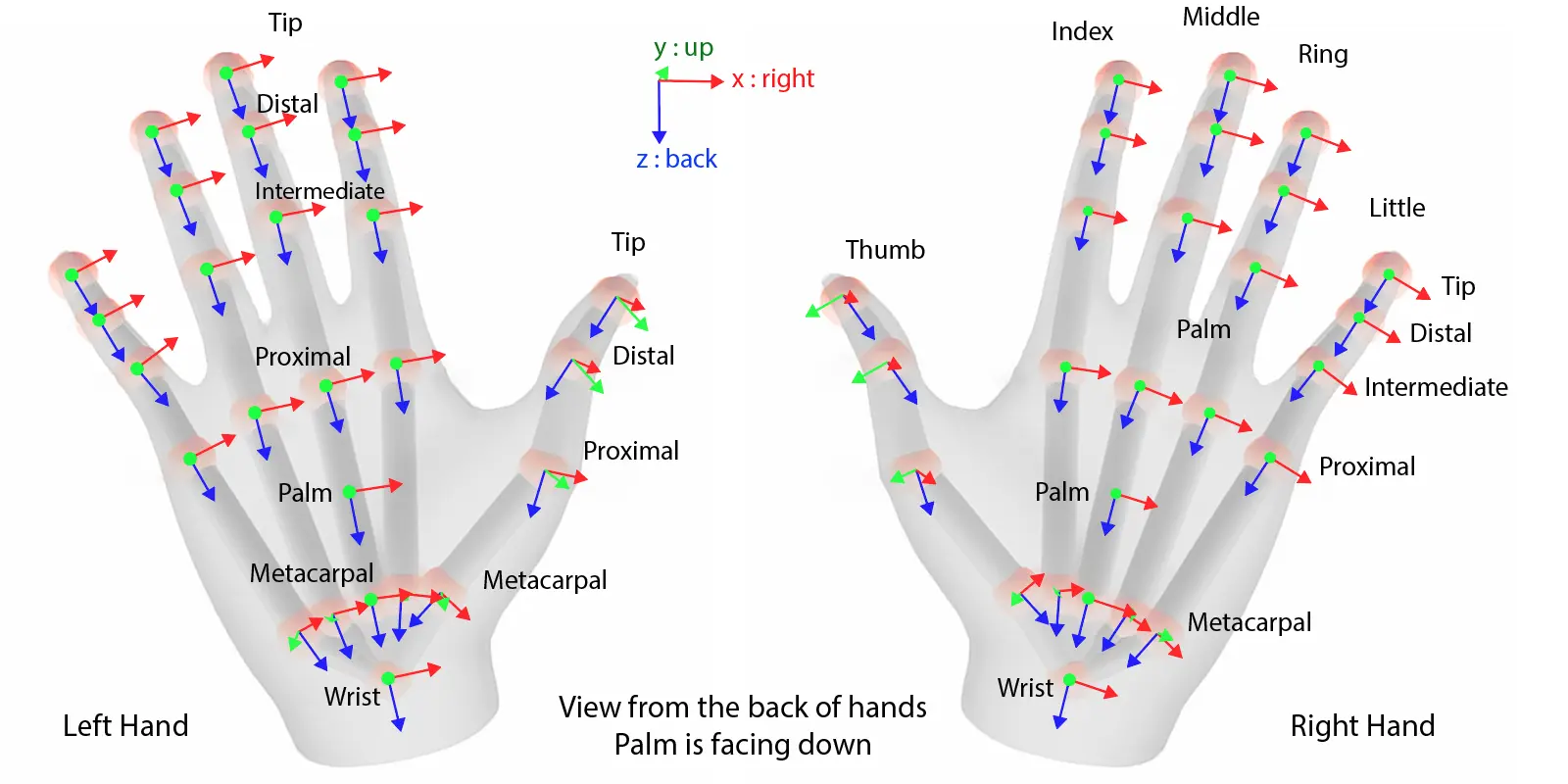}
    \caption{Visualization of the 24 OpenXR hand keypoints captured by Meta Quest 3S at 90\,Hz, providing 6-DoF wrist and finger landmarks as input for BEAVR’s coordinate transformations and control solvers.}
    \label{fig:openxr_keypoints}
\end{figure}

Hand tracking is performed using the Meta Quest 3S headset at 90\,Hz, yielding 24 keypoints per hand compliant with the OpenXR standard (see Fig.~\ref{fig:openxr_keypoints}). The resulting pose vectors serve as inputs to coordinate transformation modules that align the VR-tracked data with robot-specific frames, accommodating both dexterous hands and other end-effector types. This modular architecture enables integration with new embodiments by updating the relevant transformation matrices and solver targets without reengineering the full pipeline.

\subsection{Coordinate Transformations}

BEAVR computes the relative motion between the VR-tracked hand and the robot end-effector using homogeneous transformations. First, VR keypoints are translated to the wrist frame:
\[
P'_i = P_i - P_0,
\]
where \(P_0\) is the wrist keypoint. An orthogonal hand basis is constructed using index, middle, and pinky landmarks, followed by Gram-Schmidt orthogonalization to ensure a stable frame. 

To generalize across end-effectors, transformations are modularized:
\[
H_{RT,RH} = H_{RI,RH} \cdot (H_{R,V}^{-1} H_{HT,HI} H_{R,V}),
\]
where \(H_{HT,HI}\) encodes the VR hand’s relative motion, and \(H_{RI,RH}\) the robot's initial pose. Scale-adjusted translations are applied separately depending on the embodiment (e.g., hand vs. gripper).

This modular design allows seamless integration of new robotic embodiments by updating only the transformation matrices and solver targets.

\subsection{Inverse Kinematics Solver, Anti-Collision, and Temporal Smoothing}

We implemented a multi-target inverse kinematics (IK) approach to optimize eight fingertip positions (two per finger, excluding the pinky) and compute joint configurations for dexterous manipulation. For each fingertip position \( p \), we apply axis reflection and coordinate conversion from the Y-up VR frame to the Z-up robot frame:
\[
p' = [-p_x,\ -p_z,\ p_y],
\quad p'' = [p'_y \cdot s_f,\ -p'_x \cdot s_f,\ p'_z \cdot s_f],
\]
where the scaling factor \( s_f \) is defined as:
\[
s_f =
\begin{cases}
1.8 & \text{for index, middle, and ring fingers}, \\
1.7 & \text{for the thumb},
\end{cases}
\]
following empirical tuning informed by prior estimates of robotic hand-to-human hand scaling (\(\approx 1.6\times\)) reported in LeapHand~\cite{shaw2023leaphandlowcostefficient}.

The IK problem is solved using a Damped Least Squares (DLS) formulation, leveraging the previous joint configuration as a seed to improve stability and convergence. A dedicated anti-collision routine ensures that all computed joint solutions respect mechanical and kinematic limits, particularly under high-density or contact-rich manipulation, by integrating real-time collision checks into the IK optimization loop.

To ensure smooth, low-latency control, we apply temporal filtering on joint angles and end-effector poses. Moving average filters are applied to keypoint trajectories, while complementary filters with quaternion SLERP~\cite{10.1145/325165.325242} blending are used for pose states, mitigating jitter and producing stable control signals.

\subsection{System Integration and Communication}
Communication between the VR system, robot, and control modules is handled via ZMQ, combining publisher-subscriber channels for continuous streaming with request-reply patterns for control commands. The BEAVR system relies heavily on an efficient network module designed specifically for thread-safe messaging in real-time robotic teleoperation. \textbf{Threads} allow the central processing unit (CPU) to execute multiple tasks within the same process. When operating multiple robots, each with various interacting components a reliable communication mechanism that avoids data races, system crashes, and inconsistent behavior is crucial. To address this, we developed a dedicated network API built on top of ZMQ, providing the following key functionalities:

\begin{itemize}
\item \textbf{Publishers:} Components broadcasting messages to multiple subscribers.
\item \textbf{Subscribers:} Components receiving specific messages from publishers.
\item \textbf{Thread Management:} Ensures each thread safely manages its own ZMQ socket.
\item \textbf{Reliable Delivery:} Guarantees message delivery through handshake mechanisms, addressing potential issues such as slow-joiners.
\end{itemize}

ZMQ is chosen for its ability to efficiently abstract common messaging patterns (PUB/SUB, REQ/REP, PUSH/PULL), facilitating lightweight and high-speed communication suited for real-time robotic applications. Additionally, ZMQ inherently supports thread safety by allowing individual threads to manage their own sockets. A detailed explanation of the network system may be found in Appendix \ref{appendix:network}

A complementary Unity-based application supports real-time teleoperation, receiving keypoint data over TCP ports from a dedicated daemon process. This architecture ensures uninterrupted, high-frequency data transmission and seamless integration between human operators and robotic platforms.

The main algorithmic and architectural contributions of BEAVR are:
\begin{itemize}
    \item Stable hand frame construction using multi-landmark bases with Gram-Schmidt orthogonalization.
    \item Explicit, modular coordinate system conversion (Y-up VR to Z-up robot) via robot-specific transformation matrices.
    \item Multi-target IK for dexterous robotic hands, incorporating anti-collision routines and temporal smoothing.
    \item Temporal filtering combining moving averages and complementary SLERP blending for smooth control.
    \item Flexible, modular pipeline design enabling adaptation to diverse robotic embodiments and control frameworks.
\end{itemize}
.

\section{LeRobot}\label{sec:dataset}

The integration with LeRobot provides a standardized dataset format, models, and tools for robotic hardware. 

\subsection{LeRobot Dataset Format}
The LeRobotDataset format provides a simple yet flexible structure for robotics. It integrates with Hugging Face hub and PyTorch, allowing users to load datasets from the Hugging Face hub or a local directory. This encourages collaboration within the robotics community. Each indexed frame in a dataset includes PyTorch tensors representing observations and actions, facilitating direct usage for model training. A unique feature of the format is the ability to query temporally related frames using a delta timestamps parameter. This enables users to efficiently retrieve sequences of observations around a given frame index. Internally, the dataset is stored using widely adopted formats: observations and actions stored as Arrow/Parquet tables via the Hugging Face datasets library. Video files are compressed to MP4 files and metadata is managed using standard JSON formats. This approach ensures ease of use and extensibility for various robotics sensory inputs and state data, a practical feature for diverse robotics and learning scenarios.

\subsection{Real-time Data Streaming}
The existing LeRobot framework already contains a control and recording loop designed to handle various robotic platforms. Our teleoperation stack integrates directly into this loop, enhancing it with real-time data streaming capabilities. Specifically, this integration includes two critical functionalities:

\begin{enumerate}
\item \textbf{Capturing Observations:} The loop captures the current state of the robot and all associated sensory inputs, including real-time images from RGB and depth cameras, and joint positions.
\item \textbf{Sending Actions:} Actions are generated either through direct human teleoperation or policy inference, which are then published to the robotic system.
\end{enumerate}

Both functions employ the network module, enabling low-latency and high-frequency data exchanges.

\section{Experimental Evaluation}\label{sec:experiments}

Our experiments aim to answer the following questions.
\begin{enumerate}
    \item What range of robotics tasks can BEAVR accomplish?
    \item How well can the entire end-to-end policy training and evaluation pipeline perform on a given task?
    \item How well does the BEAVR system scale to several robots?
\end{enumerate}

We provide baseline comparisons against other VR teleoperation systems

\subsection{Experimental setup}
\begin{figure}[tb]
    \centering
    \includegraphics[width=0.9\linewidth]{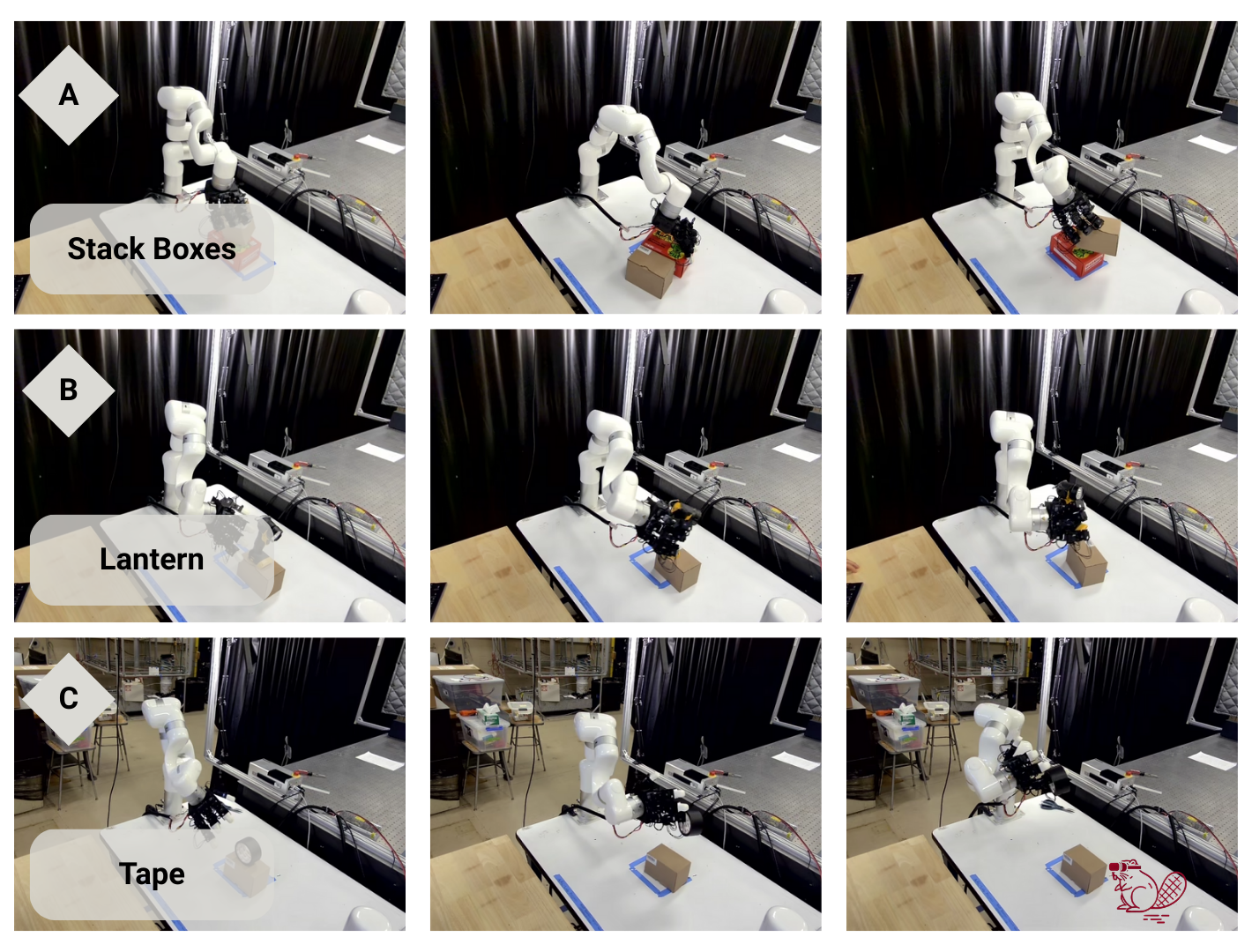}
    \caption{Benchmark manipulation tasks in our experimental setup: (A) Stack Boxes, (B) Lantern, and (C) Tape. Each row shows a chronological progression (left → right).}
    \label{fig:experimental_setup}
\end{figure}

All experiments were conducted on an Alienware x16 R2 laptop equipped with an Intel Core Ultra 9 185H CPU and an NVIDIA GeForce RTX 4080 Max-Q GPU (12GB VRAM). The system runs Ubuntu 24.04.2 LTS with CUDA 12.8 and NVIDIA driver version 570.169. Figure \ref{fig:experimental_setup} shows a sequence of images for each task, starting with the initial pose of the fixed-base 7-DoF tabletop manipulator XArm7 robot with a 16-DoF LEAP hand attachment to the end effector. All trials were conducted on a tabletop workspace. Two statically mounted RGB cameras (front-facing and overhead) captured synchronized streams at 480$\times$640 resolution and 30 FPS. We conduct three separate experiments.

\begin{enumerate}
    \item We perform three tasks demonstrating dexterous arm and hand use.
    \item We evaluate the success rate of three visuomotor policies in a given task.
    \item We analyze the performance of the entire system and its ability to scale by collecting latency, jitter, and frequency data.
\end{enumerate}

\begin{table*}[!t]
  \caption{Study comparing BEAVR expert success rate and median completion time with other baselines}
  \label{tab:expert_comparison}
  \centering
  \begin{tabularx}{\textwidth}{@{}l *{4}{Y} *{4}{Y} @{}}
    \toprule
    \multirow{2}{*}{\textbf{Task}}
      & \multicolumn{4}{c}{\textbf{Success Rate}}
      & \multicolumn{4}{c}{\textbf{Median completion time (s)}} \\
    \cmidrule(lr){2-5}\cmidrule(lr){6-9}
      & \textbf{Holo‑Dex (n=5)} & \textbf{Any‑Teleop (n=5)} & \textbf{OpenTeach (n=5)} & \textbf{BEAVR (n=5)}
      & \textbf{Holo‑Dex (n=5)} & \textbf{Any‑Teleop (n=5)} & \textbf{OpenTeach (n=5)} & \textbf{BEAVR (n=5)} \\
    \midrule
    Flip cube      & 1   & 1    & 1.0  & 1.0
                   & 6.58 & 13.71 & 2.85  & 13.32 \\
    Pour           & - & -  & 0.8  & 0.8
                   & -  & -   & 14.83 & 28.92 \\
    Pick and Place & - & -  & 0.8  & 1.0
                   & -  & -   & 11.88 & 9.72  \\
    \bottomrule
  \end{tabularx}
\end{table*}

\subsection{Task Evaluation}

We evaluate our system on six representative manipulation tasks: manipulating adhesive tape (“Tape Task”), grasping and operating a lantern (“Lantern Task”), stacking blocks of varying sizes (“Stack Blocks”), flipping a cube ("flip cube"), pouring a cup into another cup ("pour"), and a pick and place task ("pick and place"). Each task is designed to highlight different aspects of dexterous, finger-level, and arm teleoperation. For each task, a series of 10 trials is performed using our system. We record success and failure according to task-specific completion criteria, enabling quantification of task performance. The results for each task can be seen in Table \ref{tab:tasks}. The completion criteria for each task are enumerated below. We also collect five additional samples of an expert completing three tasks to compare with a baseline from other teleoperation systems. The results are found in Table \ref{tab:expert_comparison}. 

\begin{enumerate}
    \item \textbf{Tape task}: Insert the ring finger of the hand into the roll of tape, lift it and set it down on top of a box.
    \item \textbf{Lantern task}: Grasp the lantern as if to use it for illumination and place it on top of the box.
    \item \textbf{Stack blocks}: Organize two blocks into the target area by stacking them from smallest to largest (smallest on top then largest below).
    \item \textbf{Flip cube}: Successfully flip a cube to another of its sides.
    \item \textbf{Pour}: Grasp and tilt a paper cup as if to pour liquid into another paper cup.
    \item \textbf{Pick and place}: Pick up an item and place it in a target area. 
\end{enumerate}

\begin{table}[tb]
\small
\centering
\caption{Success rates across tasks}
\label{tab:tasks}
\begin{tabularx}{\columnwidth}{@{}lYY@{}}
\toprule
\textbf{Task} & \textbf{Success rate} & \textbf{Avg. time (s)} \\
\midrule
Tape task      & 6 / 10 (60\%) & 17.61 \\
Lantern task   & 8 / 10 (80\%) & 21.61 \\
Stack blocks   & 7 / 10 (70\%) & 25.89 \\
Flip cube      & 10 / 10 (100\%) & 16.50 \\
Pour           & 8 / 10 (80\%) & 84.67 \\
Pick and place & 10 / 10 (100\%) & 11.90 \\
\bottomrule
\end{tabularx}
\end{table}

The task success rate and median completion time for the other systems were not collected by us \footnote{OpenTeach numbers copied from Table IV of Iyer et al. 2024. No published numbers available for Holo-Dex and Any-Teleop for pour and pick and place tasks}, but serve as baselines for relative comparison. The "flip cube" task for the other systems was conducted in simulation while we perform a similar task on a real robot, hence, a longer completion time. Our pour task was also modified to use a LEAP hand instead of a gripper and a cup instead of sprinkles, which also explains the increased completion time.

\subsection{Policy Learning Evaluation}
For autonomous control via policy rollouts, we collect a dataset with 50 expert demonstrations using the BEAVR system, and use this dataset to train the following visuomotor policies: Action-Chunking Transformers (ACT) \cite{zhao2023learningfinegrainedbimanualmanipulation}, Action Diffusion \cite{chi2024diffusionpolicy}, and a fine-tuned version of SmolVLA \cite{shukor2025smolvlavisionlanguageactionmodelaffordable}. Policy performance is assessed by executing the trained policies on the task and recording successful completion rates. The completion criterion for the pickup box task is: grasp a box, lift it, and set it down in a target area. Table \ref{tab:policies} shows the performance of three different policies, ACT, Diffusion, and SmolVLA trained on the same dataset.

\begin{table}[tb]
\small
\centering
\caption{Success rates for policy evaluation on pickup box task}
\label{tab:policies}
\begin{tabularx}{\columnwidth}{@{}lYY@{}}
\toprule
\textbf{Policy} & \textbf{Success rate} & \textbf{Avg. time (s)} \\
\midrule
ACT             & 10 / 10 (100\%) & 9.16 \\
Diffusion       & 8 / 10 (80\%) & 23.88 \\
SmolVLA         & 7 / 10 (70\%) & 33.84 \\
Human operator  & 10 / 10 (100\%) & 12.08 \\

\bottomrule
\end{tabularx}
\end{table}

\subsection{Network Performance Evaluation}

To evaluate how the BEAVR system scales with increasing control frequency and number of effectors, we measured latency and jitter across three control configurations. Each configuration represents a different combination of robot count and command frequency, as detailed in Table~\ref{tab:control_configs}.

\begin{table}[tb]
\small
\centering
\caption{Control configurations used to evaluate network performance.}
\label{tab:control_configs}
\begin{tabularx}{\columnwidth}{@{}lYY@{}}
\toprule
\textbf{Configuration} & \textbf{Robots} & \textbf{Frequency (Hz)} \\
\midrule
1 & XArm7, LEAP         & 30  \\
2 & XArm7, LEAP         & 90  \\
3 & 2 XArm7, 2 LEAP     & 30  \\
\bottomrule
\end{tabularx}
\end{table}

Table~\ref{tab:performance_results} summarizes the achieved control frequencies and measured jitter for each robot component across these configurations. The system maintains high timing fidelity, achieving over 99\% of the target frequency in all conditions with sub-millisecond jitter.

\begin{table}[tb]
\small
\centering
\caption{Real-time performance metrics for teleoperation at different command frequencies.}
\label{tab:performance_results}
\begin{tabularx}{\columnwidth}{@{}lYY@{}}
\toprule
\textbf{Configuration} & \textbf{Achieved Frequency (Hz)} & \textbf{Jitter (ms)} \\
\midrule
XArm7 (single-arm, \SI{30}{Hz})    & 29.93 & 0.90 \\
LEAP (single-arm, \SI{30}{Hz})     & 29.69 & 0.19 \\
XArm7 (bimanual, \SI{30}{Hz})      & 29.93 & 0.89 \\
LEAP (bimanual, \SI{30}{Hz})       & 29.61 & 0.21 \\
XArm7 (single-arm, \SI{90}{Hz})    & 99.18 & 0.75 \\
LEAP (single-arm, \SI{90}{Hz})     & 97.22 & 0.13 \\
\bottomrule
\end{tabularx}
\end{table}

\begin{figure}[tb]
    \centering
    \includegraphics[width=1\linewidth]{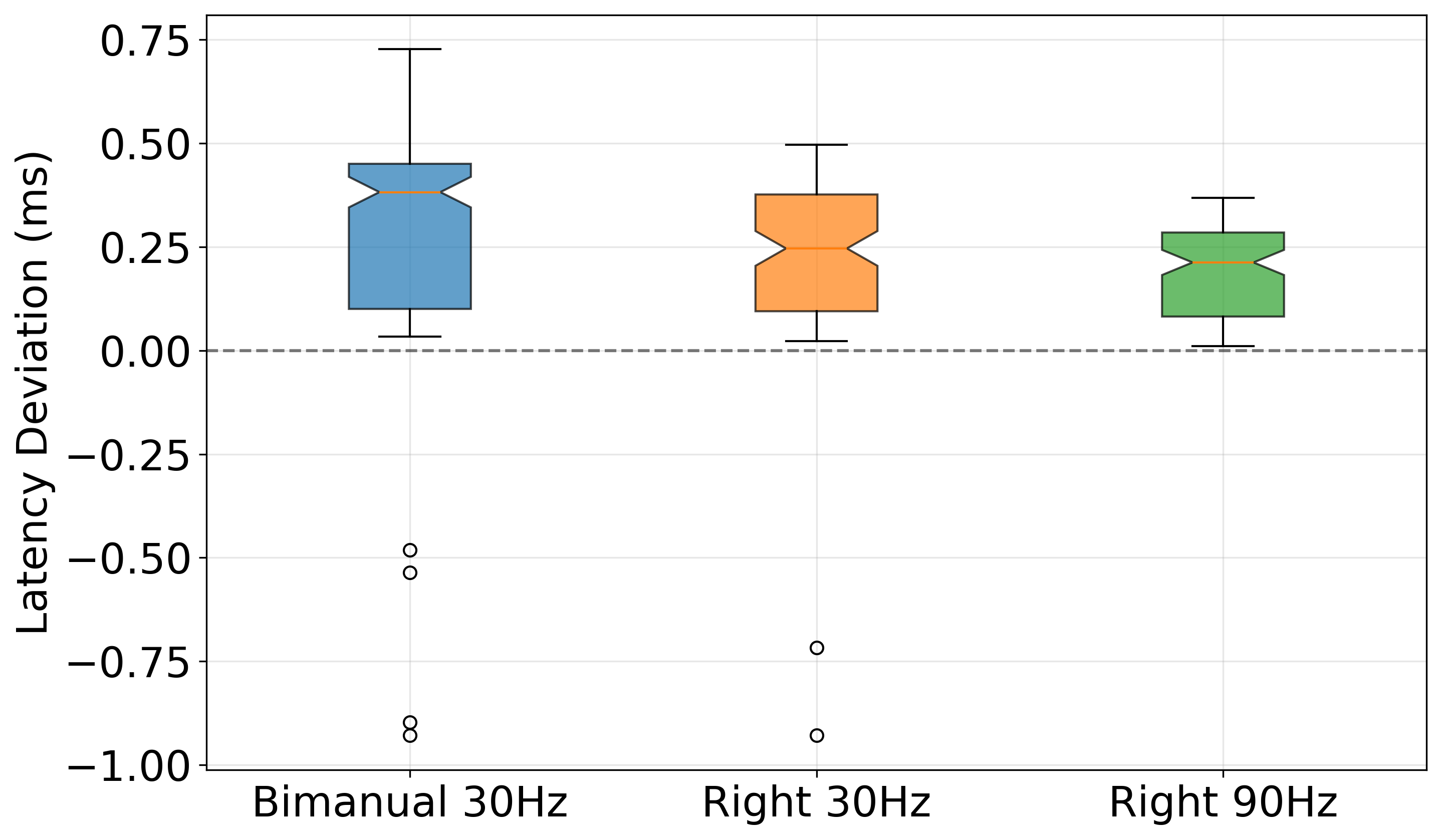}
    \caption{Latency distribution combining XArm7 and LEAP hand control. This figure aggregates arm and hand statistics into a single distribution.}
    \label{fig:latency}
\end{figure}

\begin{figure}[tb]
    \centering
    \includegraphics[width=1\linewidth]{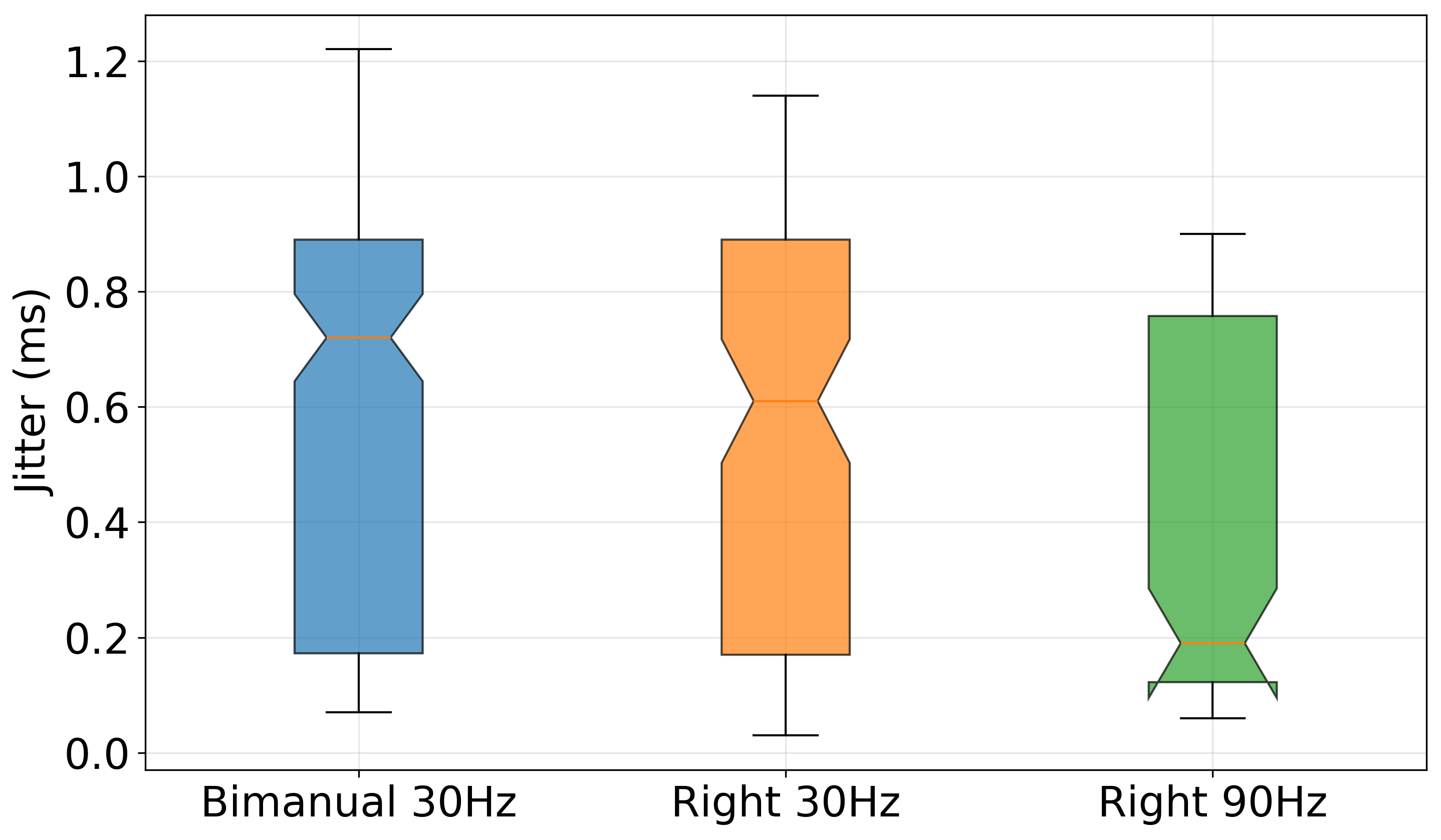}
    \caption{Jitter distribution combining XArm7 and LEAP hand control. This figure combines both arm and hand statistics into a single distribution.}
    \label{fig:jitter}
\end{figure}

Latency and jitter were recorded over 60-second control episodes and analyzed for both arms and hands. The distributions of these metrics, aggregated across all configurations, are shown in Figures~\ref{fig:jitter} and~\ref{fig:latency}. Each figure combines the statistics for both the XArm7 and LEAP across configurations, providing an overall view of temporal performance stability.

We evaluate BEAVR's network performance by benchmarking it against published metrics from teleoperation systems using diverse network architectures. At a control frequency of \SI{90}{\hertz}, BEAVR achieves approximately \SI{10}{\milli\second} of latency and sub-millisecond jitter, outperforming typical Wi-Fi-based teleoperation systems (e.g., \SI{57.4}{\milli\second} round-trip latency and 2--4~ms jitter~\cite{bray2024latency}) and VR-based interfaces such as Vicarios (\SI{40}{\milli\second} one-way latency~\cite{naceri2021vicarios}). 

At \SI{30}{\hertz}, BEAVR's latency increases to approximately \SI{33}{\milli\second}, which remains competitive, matching or surpassing other VR teleoperation platforms. While BEAVR does not yet reach the ultra-low latency of cutting-edge 5G URLLC solutions (0.8--2.0~ms latency, 1.1~ms jitter at the 99.9th percentile~\cite{reiher2022enabling}) or wired LAN-based systems (\SI{5.7}{\milli\second} round-trip latency, \SI{1}{\milli\second} jitter~\cite{bray2024latency}), it demonstrates consistent, reliable timing control even with multiple effectors operating simultaneously.

\begin{table*}[!t]
  \caption{Comparison of teleoperation system performance metrics}
  \label{tab:network_comparison}
  \centering
  \small
  \begin{tabularx}{\textwidth}{@{}lYYY@{}}
    \toprule
    \textbf{System} & \textbf{Control Frequency (Hz)} & \textbf{Latency (ms)} & \textbf{Jitter (ms)} \\
    \midrule
    \textbf{BEAVR (\SI{90}{Hz}, single-arm)} & 97-–99 & 10.1 (one-way) & 0.13--0.75 \\
    \textbf{BEAVR (\SI{30}{Hz}, single/bimanual)} & 29--30 & 33.4--33.8 (one-way) & 0.19--0.90 \\
    LAN-based teleop (Bray et al.) & 30--60 & 5.7 (RTT) & ~1.0 \\
    Wi-Fi-based teleop (Bray et al.) & 30--60 & 57.4 (RTT) & ~2.0--4.0 \\
    5G URLLC industrial (Reiher et al.) & 100+ & 0.8--2.0 & ~1.1 (99.9th \%) \\
    VR UR5 teleop (Vicarios) & ~30 & 40 (one-way) & – \\
    \bottomrule
  \end{tabularx}
\end{table*}

Across all configurations, BEAVR achieved control frequencies within $99.2\%\text{--}99.4\%$ of target, with negligible degradation when scaling to four robots or increasing command rates to \SI{90}{Hz}. Jitter remained low and consistent: under \SI{0.90}{ms} for xArm7 and under \SI{0.22}{ms} for LEAP hand in all cases. These values held steady even when doubling the number of controlled effectors. The highest-rate configuration (Config 2: \SI{90}{Hz}) exceeded expectations, with xArm7 operating at \SI{99.18}{Hz} and LEAP at \SI{97.22}{Hz}, more than three times the base \SI{30}{Hz} rate, while maintaining sub-millisecond jitter.

\section{Discussion}\label{sec:discussion}

By addressing the three experimental questions, several key findings about BEAVR’s capabilities and performance are highlighted. First, BEAVR enables dexterous finger-level manipulation across diverse tasks, including stacking, tool use, and grasping. Although success rates varied across tasks, it is important to note that all datasets were collected in single takes. Consequently, failures arose partly from user skill and partly from the mechanical dexterity of the robots themselves, rather than from limitations of BEAVR. With improved low-level control algorithms or more dexterous hardware, the range of supported tasks is expected to expand significantly.

Second, BEAVR functions as an end-to-end framework for autonomous robotic control. Teleoperation, dataset collection, policy training, and rollout evaluation are integrated into a seamless pipeline. The policy experiments demonstrate that control policies trained within this framework (e.g., ACT, DiffusionPolicy, SmolVLA) achieve strong task success rates, underscoring BEAVR’s value not merely as a teleoperation tool but also as a foundation for learning-based research.

Third, real-time performance evaluations demonstrate that BEAVR sustains reliable control across both single-arm and bimanual configurations at various command frequencies. The system consistently achieved near-target control rates (e.g., 29.93 Hz at \SI{30}{Hz}, 99.18 Hz at \SI{90}{Hz}) with low jitter, even under increased robot count and command frequency. Notably, higher-frequency control yielded lower jitter, which fell from \SI{0.90}{ms} to \SI{0.75}{ms} when we moved from \SI{30}{Hz} to \SI{90}{Hz}, indicating that the network layer and threading architecture efficiently handle real-time demands and may even improve under higher loads. Furthermore, the similarity in jitter between single-arm and bimanual setups suggests that bottlenecks are localized within local control loops rather than within the BEAVR architecture itself.

Together, these findings highlight two core strengths of the BEAVR system: its inherent flexibility in supporting heterogeneous robot platforms without sacrificing communication performance, and its robust scalability when increasing operational complexity. These properties position BEAVR as a promising open-source teleoperation platform for a broad range of research scenarios, from dexterous manipulation to multi-robot coordination, and as a valuable bridge between teleoperation and data-driven policy learning.

\section{Limitations and Future Work}\label{sec:limitationsandfutureWork}

While BEAVR demonstrates strong performance across diverse teleoperation tasks, several limitations remain that suggest future avenues for research and development.

\textbf{Hardware limitations.} One core challenge is the lack of affordable, high-performance robotic hands. The LEAP hand used in our experiments provides an accessible starting point, but its utility is currently constrained by factors such as its ability to execute precise, repeatable grasps like touching the thumb to any finger. Additionally, the physical bulk of the hand can obstruct VR keypoint tracking, particularly when the palm is oriented downward or during fist-like grasps.

\textbf{System usability and extensibility.} While BEAVR supports multiple robot embodiments, adding new hardware still requires manual configuration, calibration, and tuning. Simplifying these processes through auto-configuration tools, standardized robot templates, or GUI-based setup wizards could make the system more accessible to broader audiences.

\textbf{Call for community contribution.} As an open-source project, BEAVR depends on ongoing community engagement. We invite researchers and developers to contribute new robot modules, improve IK and retargeting components, simplify launch processes, and enhance documentation. Contributions of benchmark datasets, plugins (e.g., haptics or AR overlays), or alternate control strategies (e.g., shared autonomy) would expand the system’s impact. A more diverse ecosystem of robots and tasks would also help establish common evaluation standards for VR-based teleoperation.

\section{Conclusion}\label{sec:conclusion}

BEAVR addresses long-standing challenges in robotic teleoperation of scalability, hardware flexibility, and ease of integration by providing an open-source, modular framework that unifies control, demonstration, and learning. Its plug-and-play operator interface, built on a component-based architecture, enables real-time streaming and control at up to \SI{90}{Hz} across multiple heterogeneous embodiments, including RX1, xArm7, and LEAP hands. By natively integrating with Hugging Face’s LeRobot ecosystem, BEAVR facilitates efficient, standardized dataset collection and seamless transitions from teleoperation to policy learning. The system supports an asynchronous ``think-act'' control loop that ensures uninterrupted operation, even under high computational loads.

BEAVR’s architecture demonstrates excellent performance in both single-arm and bimanual configurations, maintaining low-latency communication (sub-\SI{35}{ms} RTT-equivalent) without requiring specialized networking infrastructure. As an open-source initiative, BEAVR aims to lower the barrier to entry for robotics research by democratizing access to high-quality teleoperation and policy learning tools.

\section{Acknowledgements}

Research was sponsored by the Department of the Air Force Artificial Intelligence Accelerator and was accomplished under Cooperative Agreement Number FA8750-19-2-1000. The views and conclusions contained in this document are those of the authors and should not be interpreted as representing the official policies, either expressed or implied, of the Department of the Air Force or the U.S. Government. The U.S. Government is authorized to reproduce and distribute reprints for Government purposes notwithstanding any copyright notation herein.

\newpage
\begin{appendices}
\section{Network Appendix}
\label{appendix:network}
The BEAVR network module comprises several key components motivated by the following concepts.
\textbf{Context}: A ZMQ context manages all sockets within a given process. This context coordinates the creation and management of sockets, I/O threads, and shared buffers, ensuring efficient resource handling and preventing interference across processes. \textbf{Sockets} are a way of connecting two machines so they can communicate. Each ZMQ socket is initialized through our API with appropriate configurations, such as high-water mark settings to limit buffering (a region of memory used to temporarily store data) and discard older messages if necessary. Subscribers run dedicated threads, continuously polling for new messages on their subscribed topics. The API also provides convenient wrappers to seamlessly serialize and deserialize Python objects alongside topics. A singleton publisher manager ensures that each unique host-port combination has exactly one dedicated publisher thread. These threads independently handle serialization, queuing, and message broadcasting, effectively isolating network operations from the main application logic. This design reduces the risk of race conditions.

To mitigate the slow-joiner issue inherent in PUB/SUB architectures, a handshake coordinator confirms the successful delivery of critical messages to subscribers. This guarantees subscribers are synchronized and ready before important messages are sent, ensuring reliable and orderly communication.

When teleoperation initiates, the system spawns individual processes for its primary components (e.g., detector, operator, interface). Each process independently creates its own sockets but shares a global ZMQ context. This setup enables immediate PUB/SUB communication across processes, promoting modularity and scalability within the teleoperation system.

\usetikzlibrary{
  positioning,      
  shapes.multipart, 
  arrows.meta       
}

\tikzset{
  process/.style={
    rectangle split,
    rectangle split parts=2,
    rectangle split part align={center,left},  
    draw,
    rounded corners,
    text width=8.5cm,
    minimum height=2.8cm,
    font=\ttfamily\footnotesize,
    fill=green!10,
  },
  arrow/.style={-{Stealth}, thick},
}

\begin{figure}[ht]
\centering
\begin{tikzpicture}[node distance=0.5cm]
  \node[process, fill=blue!20] (detector) {%
    \centering\textbf{DETECTOR PROCESS}\par
    \nodepart{second}%
    \raggedright
    ZMQ context: detector context \\
    Threads: \\
    \begin{itemize}
        \item Main loop (frame grabbing)
        \item PUB thread → tcp://*:9000 (topics: right, button, pause)
    \end{itemize}
  };

  \node[process, below=of detector, fill=red!20] (operator) {%
    \centering\textbf{OPERATOR PROCESS}\par
    \nodepart{second}%
    \raggedright
    ZMQ context: operator context \\
    Threads: \\
    \begin{itemize}
        \item Main control loop
        \item PUB thread → tcp://*:10008 (topic: endeff\_coords)
        \item SUB thread ← tcp://localhost:9002 (topic: TRANSFORMED\_*)
        \item REP Handshake server
    \end{itemize}
  };

  \node[process, below=of operator] (robot) {%
    \centering\textbf{ROBOT INTERFACE PROCESS}\par
    \nodepart{second}%
    \raggedright
    ZMQ context: interface context \\
    Threads: \\
    \begin{itemize}
        \item Main control loop (motion)
        \item PUB thread → tcp://*:10010 (topic: robot\_state)
        \item REP Handshake server
    \end{itemize}
  };

  \draw[arrow] (detector.south) -- (operator.north);
  \draw[arrow] (operator.south) -- (robot.north);
\end{tikzpicture}
\caption{Network architecture. Each component (detector, operator, interface) starts one process. Within that process, for every PUB/SUB object, we create a thread, which owns a socket. Topics support multiple logical channels (pose, buttons, commands) to flow over a single physical port}
\label{fig:teleop_network}
\end{figure}
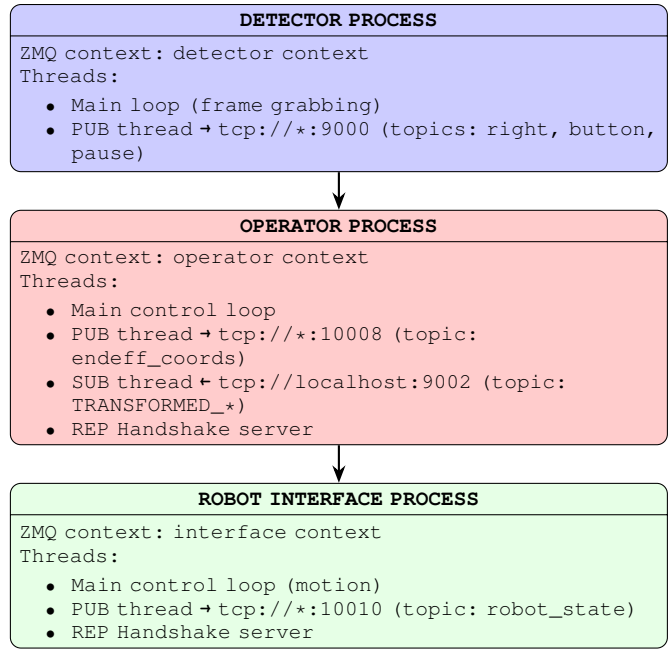

\section{Dataset Access Links}
\label{appendix:datasets}

We provide public access to the datasets collected in our experiments to encourage reproducibility and further research. The following Hugging Face URLs host the datasets for each task:

\begin{itemize}
\item \textbf{Tape Insert Dataset:} \url{https://huggingface.co/datasets/arclabmit/lx7r_tape_insert_dataset}
\item \textbf{Lantern Grasp Dataset:} \url{https://huggingface.co/datasets/arclabmit/lx7r_lantern_grasp_dataset}
\item \textbf{Stack Blocks Dataset:} \url{https://huggingface.co/datasets/arclabmit/lx7r_stack_blocks_dataset}
\item \textbf{Flip Cube Dataset:} \url{https://huggingface.co/datasets/arclabmit/lx7r_flip_cube_dataset}
\item \textbf{Pour Dataset:} \url{https://huggingface.co/datasets/arclabmit/lx7r_pour_dataset}
\end{itemize}

\end{appendices}

\vspace{10mm}

\bibliographystyle{IEEEtran}
\bibliography{references}

\end{document}